\documentclass[conference]{IEEEtran}
\IEEEoverridecommandlockouts
\usepackage{cite}
\usepackage{amsmath,amssymb,amsfonts}
\usepackage{algorithmic}
\usepackage{graphicx}
\graphicspath{{images/}}
\usepackage{textcomp}
\usepackage{xcolor}
\usepackage{cleveref}
\usepackage{url}
\usepackage{booktabs}
\usepackage{makecell}
\usepackage{flushend} 

\usepackage{pgfplots}
\pgfplotsset{compat=1.16}

\usepackage{todonotes}
\def\BibTeX{{\rm B\kern-.05em{\sc i\kern-.025em b}\kern-.08em
    T\kern-.1667em\lower.7ex\hbox{E}\kern-.125emX}}
    
\newcommand{\q}[1]{``#1''}

\makeatletter
\def\ps@IEEEtitlepagestyle{%
  \def\@oddfoot{\mycopyrightnotice}%
  \def\@oddhead{\hbox{}\@IEEEheaderstyle\leftmark\hfil\thepage}\relax
  \def\@evenhead{\@IEEEheaderstyle\thepage\hfil\leftmark\hbox{}}\relax
  \def\@evenfoot{}%
}
\def\mycopyrightnotice{%
  \begin{minipage}{\textwidth}
  \centering \scriptsize
  Copyright~\copyright~2023 IEEE. Personal use of this material is permitted. Permission from IEEE must be obtained for all other uses, in any current or future media, including\\
  reprinting/republishing this material for advertising or promotional purposes, creating new collective works, for resale or redistribution to servers or lists, or reuse of any copyrighted component of this work in other works.
  \end{minipage}
}
\makeatother

\begin{document}
\bstctlcite{IEEEexample:BSTcontrol}

\title{A Comprehensive Review on Ontologies for Scenario-based Testing in the Context of Autonomous Driving}

\author{

Maximilian~Zipfl$^{1,2,\dagger}$,
Nina~Koch$^{2,\dagger}$, 
and J.~Marius~Zöllner$^{1,2}$
\thanks{$^{1}$FZI Research Center for Information Technology, Karlsruhe, Germany
{\tt\small \{zipfl, zoellner\}@fzi.de}}%
\thanks{$^{2}$KIT Karlsruhe Institute of Technology, Karlsruhe, Germany}
\thanks{$^{\dagger}$ Both authors contributed equally to this work as first authors.}
}%

\maketitle

\begin{abstract}
The verification and validation of autonomous driving vehicles remains a major challenge due to the high complexity of autonomous driving functions. Scenario-based testing is a promising method for validating such a complex system.
Ontologies can be utilized to produce test scenarios that are both meaningful and relevant. One crucial aspect of this process is selecting the appropriate method for describing the entities involved. The level of detail and specific entity classes required will vary depending on the system being tested. It is important to choose an ontology that properly reflects these needs.

This paper summarizes key representative ontologies for scenario-based testing and related use cases in the field of autonomous driving.
The considered ontologies are classified according to their level of detail for both static facts and dynamic aspects. Furthermore, the ontologies are evaluated based on the presence of important entity classes and the relations between them. 
\end{abstract}


\section{Introduction}
\label{sec:introduction}

Autonomous Driving has become one of the most important strategic research areas in the automotive industry by promising to enhance the driving safety and offering more comfort to the drivers. A common approach to define automated driving capabilities is to differentiate between five levels of driving automation. For the fifth level, full driving automation, the Operational Design Domain (ODD), which specifies the operating conditions under which the driving automation system is designed to function, is unlimited.

One of the main challenges for full driving automation is the validation of the autonomous driving system across all relevant driving scenarios of the ODD to ensure that it behaves as intended. 
A model to describe or specify the ODD can be, for example, the 6-layer-Model from Scholtes et al. \cite{scholtes_6-layer_2021}.
Scenario-based validation is one technique for this purpose. According to Ulbrich et al. \cite{ulbrich_defining_2015} a scenario is \q{the temporal development between several scenes in a sequence}, where the transition between the scenes is expressed by actions, events and goals. Each scene is described by a \q{snapshot of the environment including the scenery and dynamic elements, as well as all actors’ and observers’ self-representations and relationships}.

Highly automated vehicles can be tested using various methods, one of the most promising being scenario-based validation \cite{stellet_testing_2015}.
Scenario-based validation systematically derives a set of critical scenarios based on a data-driven or knowledge-driven approach. The data-driven approach takes real world data from road tests, accident data, field tests and simulation. The data is often enriched by additional information, such as criticality metrics like Time-to-Collision (TTC), which is then clustered into scenarios by applying machine learning techniques. Another interesting approach in the context of data-driven scenario generation is for example demonstrated by Feng et al. \cite{feng_intelligent_2021}. Their idea is to build and use an intelligent, simulated driving environment, where background vehicles are trained via reinforcement learning to execute small, adversarial maneuvers. Further researches on the data-driven approach are outlined by Elgharbawy et al. \cite{elgharbawy_ontology-based_2019}, Ding et al. \cite{ding_cmts_2019} and Langer et al. \cite{langner_estimating_2018}.

Contrary, the knowledge-driven approach takes knowledge from existing scenarios, functional descriptions, experts, as well as from traffic guidelines and represents it in an ontology. The achieved knowledge model of the ODD serves as an input model for combinatorial algorithms to generate a large suite of test scenarios \cite{riedmaier_survey_2020, wotawa_ontology-based_2020, klueck_using_2018}. The test scenarios are afterwards stored in a database, concretized and reduced to a set of critical test scenarios based on criticality metrics \cite{leitner_enable-s3testing_2019, butz_virtual_2020}.

There are several approaches proposing ontologies for knowledge-driven scenario-based validation, such as \cite{klueck_using_2018, wotawa_ontology-based_2020, bagschik_ontology_2018, chen_ontology-based_2018}. However, there exists only a limited amount of reviews on the application of ontologies for scenario-based validation and other research areas in autonomous driving. The surveys of Nalic et al. \cite{nalic_scenario_2020} and Riedmaier et al. \cite{riedmaier_survey_2020} for example focus mainly on the classification and terminology of scenarios and scenario generation methods. They don't provide a review and categorization of ontologies. Schäfer et al. \cite{schafer_no_2017} concentrate on the improvements ontologies can bring to systems engineering and requirements analysis, but they leave out other application areas for autonomous driving. Brunner et al. \cite{brunner_ontologies_2017} provide a review on ontologies and their application for robotics and autonomous driving. They categorize ontologies into different application areas for autonomous robots and vehicles. However, their review discusses only a few approaches for each application area, doesn't compare the ontologies between the different application areas and doesn't classify the ontologies according to their level of detail.

Our survey provides a comprehensive overview and evaluation of ontologies for scenario-based validation and other application areas of autonomous driving. Our contributions are as followed:
\begin{itemize}
     \item \emph{Comprehensive Review} We provide a detailed overview of ontologies and research approaches for different application areas within autonomous driving.
    \item \emph{New Classification} Based on our comprehensive review, we classify the identified ontologies into two different application areas for autonomous driving: Scenario-based validation and situation assessment and decision-making.
    \item \emph{Categorization \& Evaluation} We evaluate the completeness of the ontologies by reviewing important classes and relations and comparing the findings between each ontology. In addition, we rate the level of detail of static aspects and dynamic factors that can be described by the ontologies. This categorization supports other researchers in choosing a suitable ontology that fits to their needs and validation task.
\end{itemize}

The rest of this review is structured as followed. \Cref{sec:definition} outlines the background, definition and notation of ontologies. \Cref{sec:ontologies} categorizes the ontologies into the different application areas within autonomous driving and compares them. \Cref{sec:evaluation} evaluates the completeness and compares the abstraction level of each of these ontologies. \Cref{sec:conclusion} summarizes the paper with a conclusion.

\section{Definition and Background of Ontologies}

\label{sec:definition}
The term \q{Ontology} comes from the Greek \q{ontos} for \q{being} and \q{logos} for \q{word} \cite{gasevic_dragan_model_2009}. In Computer Science, an ontology formally models the structure of a domain and can be used as a knowledge-base by an application \cite{guarino_nicola_handbook_2009, bagschik_ontology_2018}. According to Studer et al. \cite{studer_knowledge_1998} \q{an ontology is a formal, explicit specification of a shared conceptualization}. A conceptualization can be understood as an \q{abstract simplified view of the world}, as defined by Genesereth and Nilsson \cite{Genesereth01}.

Ontologies use a collection of statements. The statements define relations between concepts to describe the perceived view of a domain \cite{berners-lee_tim_semantic_2001}. Next to concepts and relations, an ontology consists of individuals, which are instances of the concepts. Attributes and attribute values further describe the concepts. Additionally, ontologies use axioms, which constrain the interpretation of the concepts, relations and instances \cite{jakus_concepts_2013, guarino_nicola_handbook_2009}.


The main benefits of ontologies can be derived from the definition of Studer et al. \cite{studer_knowledge_1998}. First, ontologies are shared and explicit. They capture common knowledge and provide unambiguous vocabulary that several users have agreed on \cite{jakus_concepts_2013, gasevic_dragan_model_2009}. Second, ontologies are formal. They are specified by a machine-readable language and can be processed by computers, thus facilitating interoperability and reusability \cite{lv_approach_2011}. Third, ontologies rank concepts and relations hierarchically in a taxonomy. It is possible to infer knowledge upon an ontology due to the semantic relations and rules that can be defined \cite{fensel_web_2008}.



\section{Ontologies}
\label{sec:ontologies}
In the context of autonomous driving, ontologies are used in two main application areas. On the one hand, ontologies are used to generate scenarios for scenario-based testing. On the other hand, ontologies are applied to evaluate the perceived environment of the automated vehicle, assess the situation and decide based on these results on the next actions.
In the following, representative ontologies are summarized for these two application areas.

\subsection{Scenario-based Validation}
Scenario-based validation uses ontologies as part of the knowledge-driven approach to represent the ODD and generate test scenarios by the application of combinatorial algorithms. Klück et al. \cite{klueck_using_2018} and Wotawa et al. \cite{wotawa_ontology-based_2020} represent relevant influence factors and test parameters in an ontology. They transform the ontology into an input model for a combinatorial algorithm to generate a test suite of abstract scenarios. The abstract scenarios are concretized based on n-wise parameter combination and a machine-learning approach. Their ontology covers the road infrastructure including slope, surface and lane, as well as the ego vehicle with its position and speed.

Similarly, Bagschick et al. \cite{bagschik_ontology_2018} construct an ontology which models the five layers of a scenario. These are the road, the traffic infrastructure, the temporary manipulation of the road and traffic infrastructure for example through construction sites, the static and dynamic objects and the environment layer. They define the degree of complexity, the amount of positions per lane, the amount of traffic participants and the abstraction level before they use combinatorial techniques and permutations to generate a start and an end scene of a functional scenario\footnote{Functional scenarios are described by natural language}. Afterwards, they transform the functional scenarios into logical scenarios\footnote{Logical scenarios are described by parameter ranges and state spaces. Both logical and functional scenarios always describe a set of possible concrete scenarios.} by adding actions and events, specifying parameter ranges, as well as object and parameter dependencies \cite{menzel_functional_2019}.

Chen and Kloul \cite{kloul_advanced_2020} use a three layered methodology for the generation of test scenarios. The first layer consists of a highway, a weather and a vehicle ontology to model the static and mobile concepts of a scene. The second layer describes the interactions between the static and mobile concepts by the use of rules expressed in first-order-logic. The third layer is the generation layer, which adds several scenes to a scenario by considering actions and events.

The simulation scenario generation framework of Medrano-Berumen and Akbas \cite{medrano-berumen_abstract_2019} uses a matrix where each row represents a road piece or actor in a semantic string. To define a road piece the string uses parameters such as road piece type, road length, number of lanes, speed limit and intersection pattern. For the description of actors, parameters, such as actor type, path type, moving speed, start location and offset are included. Based on the semantic matrix and user's input different scenarios are randomly generated and the ego vehicle is placed in them. The simulation framework collects the test data and evaluates each run. Environmental conditions are not considered, as they don't aim at testing the full vehicle stack and focus only on testing the decision component of an autonomous vehicle \cite{medrano-berumen_abstract_2019}.

Feilhauer \cite{feilhauer_absicherung_2018} on the other hand uses equivalence classes to generate concrete test scenarios out of an ontology which models the scenario components and their parameters. Equivalence classes represent groups of parameter input ranges, where the behaviour of a system doesn’t change within each of these classes. Through the combination of representatives of each of these groups, the concrete test scenarios can be generated. Additionally, it is possible to only consider the corner representatives of each class and generate corner test cases. A similar test generation methodology is presented by Schuldt et al. \cite{schuldt_effiziente_2013}.

Elgharbawy et al. \cite{elgharbawy_ontology-based_2019} approach scenario generation differently from knowledge-driven methods. They take a data-driven approach by using recordings from various sources such as bus system data, sensors, and videos. First, they clean the data by removing outliers and noise. Then they identify specific driving situations by analysing the time series and standard deviation of the data. They enrich the data by finding similarity between different time-series signals and use unsupervised machine learning algorithms to cluster the data into scenarios. Finally, they use data regression to identify characteristic signals for each scenario cluster that can be used to generate more scenarios using an ontology.

Herrmann et al. \cite{herrmann_using_2022} provide another example of how ontologies can be applied for a data-driven scenario generation approach. They use an ontology to create machine-processable descriptions of images showing different traffic situations in order to create further images for training and test datasets. These datasets are used for the validation of autonomous driving functions. Additionally, they use their ontology for annotating and labelling training data, as well as inferring a safety argumentation. Since the ontology is used for the generation and labelling of images, it considers concepts, such as different light sources, clothes of pedestrians, buildings, and colours next to concepts such as road condition and subjects that can be also found in other ontologies.

Another topic that is discussed in the context of scenario-based testing are corner cases. Bogdoll et al. \cite{bogdoll_one_2022} propose the use of ontologies to help create and model different corner case scenarios for autonomous vehicles. The proposed master ontology is able to model any scenario found in the research literature related to corner cases, which can be converted into the OpenSCENARIO format and tested in simulation. This allows for testing under difficult conditions that may not have been encountered during training data collection.

In the work of Westhofen et al. \cite{westhofen_using_2022} an approach, which combines a set of ontologies to enable the description of critical scenarios, is proposed.
Through the sub-ontologies included as modules, a very detailed description of the environment can be created.
It uses a 6-Layer Model as basis for the sub-ontologies to create a formal representation, which is then used in combination with an a-priori predicate augmentation and description logic/rule reasoner for executing automated analyses.



\subsection{Situation Assessment \& decision-making}
Next to scenario-based validation, ontologies are widely applied for situation assessment and decision-making of the ego vehicle. Ulbrich et al.\cite{ulbrich_graph-based_2014} aggregate information from environment perception modules, a-priori maps and C2X information in an ontology that is used for context modelling and that is queried during decision-making of the ego vehicle. Their ontology contains information on the ego vehicle, other objects, infrastructure and knowledge of the situation, such as planned routes and actions. Apart from that, their ontology includes constraints to represent traffic rules, as well as topological and metric attributes.

Next to the outlined work above, Hülsen et al. \cite{hulsen_traffic_2011} are one of the first researchers who demonstrated how ontologies can improve the decision-making of the ego-vehicle at an intersection. Using the example of a truck pulling out to turn right at an intersection, they show that it is not enough to process sensor data, but to be able to reason on them to predict future maneuvers. Therefore, their ontology includes relations such as hasToYield, hasRightOfWay, hasToStop, approachesTo, departsFrom that are reasoned by the ego vehicle.

Buechel et al. \cite{buechel_ontology-based_2017} use a similar ontology-based approach for decision-making of autonomous vehicles. However, they incorporate generic traffic rules and national traffic regulations in their ontology to facilitate the usage of the ontology across various countries. Whenever possible, traffic rules are modelled in a generic way, for example by using the relation hasConflictingConnector between the connectors of lane segments. Based on this relation, the autonomous vehicle can reason on right-of-way rules. In case it is not possible to model traffic rules in a generic way, traffic rules are represented as an instance of the class "TrafficRule". The autonomous vehicle can infer the correct traffic rule by considering the traffic situation and the region.

Regele et al. \cite{regele_using_2008} focus similarly to Hülsen et al. \cite{hulsen_traffic_2011} on the traffic coordination at intersections. However, contrary to the other researchers, he explicitly describes the challenge of choosing the right abstraction level for one's model. He distinguishes between a low-level world model for trajectory planning and a high-level world model for the traffic coordination problem. According to Regele et al. \cite{regele_using_2008} the high-level world model doesn't have to represent the detailed information, trajectory and geometry of all objects as the actual collision avoidance is happening at the lower level. The high-level world model for the traffic coordination should simply store the semantic meaning and facilitate a quick decision-making. Therefore, their high-level semantic traffic model consists of a set of lane objects, lane sections and lane relations with their own set of attributes, such as speed, acceleration or length of lane. The conflicting relation between two lane sections indicates that the two lane sections are overlapping and can lead to potential collisions. The ego vehicle uses this relation as part of a rule to plan its next actions.\\

Fang et al. \cite{fang_ontology-based_2019} recommend to use a probabilistic approach to forecast the long-term behaviour of nearby traffic participants. Based on the detection and localization results, they create the encountered situation and represent it in an ontology including the relations between the road users and their potential behavioural intention series. A Markov Logic Network is used by a behaviour reasoner to infer the probabilities of potential behavioural intentions based on the represented situation.

Contrary to Fang et al. \cite{fang_ontology-based_2019}, Huang et al. \cite{huang_ontology-based_2019} take a deterministic approach for decision-making. They propose an ontology to model the driving scene consisting of the ego vehicle, its behaviour, a driving scenario, an obstacle and a road network. Based on the input from the ontology and the information from the vehicle's sensors, they then assess the scenario in eight regions of interest surrounding the ego vehicle using criticality indicators like TTC. Additionally, they query the ontology to determine whether specific actions are permitted. A reasoner receives the results and determines the next course of action.

Zhao et al. \cite{zhao_ontology-based_2017} present a comparable strategy. The perceived sensor data is transformed into a Resource Description Framework input stream and stored in an ontology. A query engine retrieves information about the environment, position, and direction of the vehicle from the ontology. A reasoner then draws conclusions from the ontology's data and determines the subsequent actions, which are communicated to a path planning system. The used ontology is made up by three sub-ontologies, which are the map, control and car ontology. The control ontology allows describing the path of the ego vehicle by connecting multiple, successive path segments \cite{zhao_ontology-based_2017}.

Kohlhaas et al. \cite{kohlhaas_semantic_2014} collect information from sensors and use an ontology with classes such as vehicle, lane segment, intersection and trajectory to model the current traffic scene of the ego vehicle. Contrary to the other approaches, their ontology additionally models the semantic state space of the ego vehicle, i.e. all possible states that the ego vehicle can transition to over time, for high-level manoeuvre planning. A semantic state is described by the lateral and longitudinal relation of the ego vehicle to other dynamic objects and its position on the lane segment. All possible semantic states which the ego vehicle can transition to at a certain point of time build together the configuration. All possible configurations in turn form the semantic state space. The transitions between two states can hold information about its probability and validity \cite{kohlhaas_semantic_2014}.

As Fang et al. \cite{fang_ontology-based_2019} and Hülsen et al. \cite{hulsen_traffic_2011} emphasize, one of the challenges in applying ontologies in the research area of autonomous driving is the need for real-time processing. To overcome this issue, Hülsen et al. propose to structure the ontology into modules, so that only the relevant modules are loaded. This type of modular ontology is referred to as a lean ontology. Similarly, Zhao et al. \cite{zhao_ontology-based_2017} employ a sub-knowledge base to expedite reasoning time in their approach.

\section{Categorization and Evaluation}
\label{sec:evaluation}

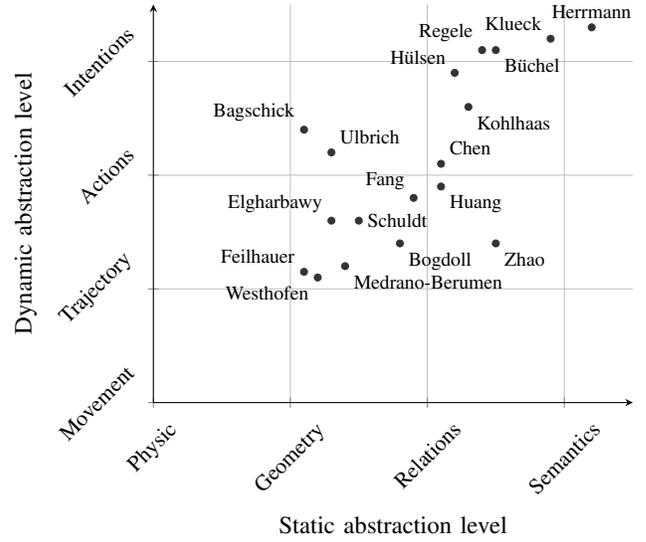
\begin{figure}[ht]
    \centering
    \scalebox{0.93}{%
    \newcommand\NodeWithText[4]{%
    \node[black,#3] at (axis cs:#1,#2){\footnotesize{#4}};
    \node[black] at (axis cs:#1,#2) [circle, scale=0.3, draw=black!80,fill=black!80] {};
}

\begin{tikzpicture}[]
\begin{axis}[
    grid = major,   
    axis x line=left,
    axis y line=left,
    xlabel={Static abstraction level},
    ylabel={Dynamic abstraction level},
    ticklabel style={rotate=45, font=\small},
    xtick={0,1,...,3},
    xticklabels={Physic, Geometry, Relations, Semantics},
    ytick={0,1,...,3},
    yticklabels={Movement, Trajectory, Actions, Intentions},
    xmin=0,
    xmax=3.5,
    ymin=0,
    ymax=3.5]
    
\NodeWithText{2.9}{3.2}{above left}{Klueck}
\NodeWithText{2.4}{3.1}{above left}{Regele}
\NodeWithText{1.1}{1.15}{above left}{Feilhauer}
\NodeWithText{1.4}{1.2}{below right}{Medrano-Berumen}
\NodeWithText{1.3}{1.6}{above left}{Elgharbawy}
\NodeWithText{1.5}{1.6}{right}{Schuldt}
\NodeWithText{2.5}{1.4}{below right}{Zhao}
\NodeWithText{1.3}{2.2}{above right}{Ulbrich}
\NodeWithText{1.1}{2.4}{above left}{Bagschick}
\NodeWithText{2.1}{2.1}{above right}{Chen}
\NodeWithText{2.1}{1.9}{below right}{Huang}
\NodeWithText{2.2}{2.9}{above left}{Hülsen}
\NodeWithText{2.3}{2.6}{below right}{Kohlhaas}
\NodeWithText{1.9}{1.8}{above left}{Fang}
\NodeWithText{2.5}{3.1}{below right}{Büchel}
\NodeWithText{1.8}{1.4}{below right}{Bogdoll}
\NodeWithText{3.2}{3.3}{above}{Herrmann}
\NodeWithText{1.2}{1.1}{below left}{Westhofen}
\end{axis}
\end{tikzpicture}}
    \caption{Space for categorizing scenario ontologies based on the static and dynamic abstraction level. Reality is located in the origin.}
    \vspace{-2ex}
    \label{fig:categorisation}
\end{figure}

 Particularly in the development process and the associated validation of autonomous vehicles, individual submodules all the way to the entire system are tested in various test methods adapted for specific purposes. In order to design the V-model-based validation strategy as efficiently as possible, the various simulation environments require different levels of abstraction. An overloaded simulation environment is development-intensive and requires more computing power to complete the test run in the same amount of time. Therefore, only as much information as necessary should be simulated. To make it more convenient to evaluate and classify the ontologies that can be used for scenario-based testing, we have categorized the ontologies described earlier into a two-dimensional overview. The categorization supports in choosing the appropriate ontology for one's validation task along the V-model and corresponding abstraction level. Since the ontologies classified in the context of situation assessment \& decision-making are based on the same components, inherently, we evaluate them alongside the ontologies of scenario-based validation.

The categorization is formed by the static abstraction axis and the dynamic abstraction axis (see \Cref{fig:categorisation}). Both axes start in the origin with the finest possible resolution and accuracy. This implies that reality lies in the origin of the coordinate system. Ontologies serve as a description model of reality and are supposed to describe it in an abstract way. An abstract description leaves room for both interpretation and action for the simulated course of the scenarios. Furthermore, a higher degree of abstraction usually leads to a more lightweight specification.

The abscissa in our categorization space describes the abstraction level of the static properties of the objects in the scenario. Thereby, the level of detail of the object decreases the further one moves to the right on the axis. The axis starts in the origin with the physical properties of an object and moves to their geometry. Geometry describes the spatial position, dimensions, metrics and curves of an object, such as the length or GPS position of the ego vehicle or the curve and width of a road. One level up on the axis are relations. Ontologies on this level focus on the relations between the objects and describe, for example, connections between road segments or potential collision risks between traffic participants. The highest static abstraction level are semantics. Ontologies on this level provide a simple, semantic overview of the objects in the scenario without describing their geometry or relations. Often, abstract ontologies with implicit information allow to derive explicit information. For example, if the surface is precisely described with physical properties, the geometry of the object can be derived from it.

On the ordinate axis, the level of detail of the dynamic movements of objects is shown. Intentions describe movements in the most abstract way. They often specify a global goal, but not how this goal is actually reached. How a goal is achieved can be described by the more precise category “Actions”. The following example shows the difference between the categorizations. The goal of a vehicle is to drive from the beginning of a road to the next intersection. With actions, it is additionally defined that the vehicle in front is to be overtaken. If we additionally specify a trajectory, both the location and the time of the overtaking maneuver are defined. In the movement category, not only the center of mass of the object is considered, but all parts connected with it. This can be relevant for example for intention recognition of pedestrians, where movements of limbs or head are also described (compare e.g. \cite{deng_skeleton_2018}). 

Most ontologies are in the range where the rough geometry of a vehicle, but mostly their relations and trajectories or actions can be described. In general, a trend can be seen that newer ontologies and V\&V methods, for example from Feilhauer \cite{feilhauer_absicherung_2018}, Medrano-Berumen \cite{medrano-berumen_abstract_2019} or Bogdoll \cite{bogdoll_one_2022}, have a higher level of detail, which could be explained by more comprehensive simulation tools and more available computing power.

The ontology with the highest level of detail in our investigation is the work of Feilhauer \cite{feilhauer_absicherung_2018}. The road geometry is described by OpenDRIVE\cite{asam_asam_nodate}, which provides an exact positional representation of a road. Static objects are mapped by their exact position in a street coordinate system. Signs, which can be unambiguously defined in their size and by standards, are directly included in the ontology here. Objects that exist in different designs (trees, poles, ...) are further defined by size and an object catalogue.
The dynamic components, i.e. the behaviour of the traffic participants, are not described by a concrete trajectory, but rather by a driving region in which the vehicle can operate, for example, in order to be able to change lanes and independently adjust the speed in a defined range.
This high level of detail is particularly important in this case, since testing of the complete trajectory planning should be enabled in a wide ODD, which can be specified by a logical scenario description.

Contrary, the ontology of Klück et al. \cite{klueck_using_2018} has a low level of detail. For example, a road is modelled by a set of n lanes that are limited by one or two lines and that have a traffic and surface condition. The surface conditions are described by abstract, semantic values such as dry, slippery or icy. Similarly, traffic conditions are defined as smooth, light, heavy or jam. Klück et al. \cite{klueck_using_2018} intentionally use an abstract, semantic description model, as their goal is to create a set of abstract, functional scenarios for scenario-based testing with the support of combinatorial algorithms. These combinatorial algorithms require a limited set of possible parameters and values to successfully work within the given computational conditions \cite{klueck_using_2018}. The functional scenarios are concretized for the use in simulations in a later step by using functions which map abstract values to real physical values.

The comparison between Feilhauer \cite{feilhauer_absicherung_2018} and Klück et al. \cite{klueck_using_2018} emphasizes that the abstraction level of an ontology depends on the autonomous driving module under test and whether functional or logical scenarios should be generated. This observation is also confirmed by considering the other papers for scenario-based validation. Klück et al. \cite{klueck_using_2018}, Bagschick et al. \cite{bagschik_ontology_2018} and Chen and Kloul \cite{kloul_advanced_2020} all propose a method for the knoweldge-driven generation of functional scenarios. Therefore, their ontologies are located at the top of the coordinate system and have a higher abstraction level. Contrary, Medrano-Berumen et al. \cite{medrano-berumen_abstract_2019}, Feilhauer \cite{feilhauer_absicherung_2018} and Schuldt et al. \cite{schuldt_effiziente_2013} generate logical scenarios. As a consequence, their ontologies are located at the bottom of the coordinate system and are more detailed.

Another observation is that most of the ontologies for situation assessment and decision-making are located in the area where relations, trajectory and actions intersect in the coordinate system. This observation makes sense, as the ontologies for this application area have to describe relations between traffic participants, resulting collision risks, as well as right-of-way rules and corresponding actions.

So far, hardly any ontologies for autonomous driving have been found, which focus on the detailed, physical level. However, when extending the search to sensor-based testing, many different approaches are discussed, especially in the radar domain \cite{magosi_survey_2022}.

\begin{table*}[ht!]
    \centering
    \caption{Representability of different categorizations by the investigated ontologies. The state of traffic participants (State of TP) is described by the position p, the velocity v and their acceleration a.}
    \begin{tabular}{lcccccccccccc}
        \toprule
        
        ~ &
    	\thead{Road\\ Geometry} &
    	\thead{Road\\ Condition} & 
    	\thead{Road \\Markings} &
    	\thead{Road \\Type} &
    	\thead{rel. \\Position} &
    	\thead{State \\of TP} &
    	\thead{Intentions/\\Actions} &
    	\thead{Traffic \\Rules} &
    	\thead{Static \\Objects} &
        Signs &
        Environment \\\midrule

        Bagschik \cite{bagschik_ontology_2018} & \checkmark & \checkmark & \checkmark & ~ & \checkmark & p & \checkmark & \checkmark & \checkmark & ~ & \checkmark \\ 
        Klück \cite{klueck_using_2018} & \checkmark & \checkmark & \checkmark & ~ & ~ & pv & ~ & ~ & ~ & ~ & ~ \\ 
        Wotawa \cite{wotawa_ontology-based_2020} & ~ & ~ & ~ & ~ & ~ & pva & Driver Type & ~ & ~ & ~ & ~ \\ 
        Chen \cite{chen_ontology-based_2018}  & \checkmark & ~ & \checkmark & \checkmark & \checkmark & pv & \checkmark & ~ & \checkmark & \checkmark & \checkmark \\ 
        Medrano-Berumen \cite{medrano-berumen_abstract_2019}  & \checkmark & ~ & ~ & \checkmark & ~ & pv & ~ & \checkmark & ~ & ~ & ~ \\ 
        Feilhauer \cite{feilhauer_absicherung_2018} & \checkmark & ~ & \checkmark & \checkmark & ~ & pva & \checkmark & ~ & \checkmark & \checkmark & \checkmark \\ 
        Schuldt \cite{schuldt_beitrag_2017} & \checkmark & ~ & \checkmark & \checkmark & \checkmark & v & ~ & ~ & \checkmark & ~ & ~ \\ 
        Elgharbawy \cite{elgharbawy_ontology-based_2019} & ~ & ~ & \checkmark & ~ & ~ & p & ~ & ~ & \checkmark & ~ & ~ \\ 
        Ulbrich \cite{ulbrich_defining_2015}& \checkmark & ~ & ~ & \checkmark & ~ & p & \checkmark & \checkmark & ~ & \checkmark & ~ \\ 
        Hülsen \cite{hulsen_traffic_2011}& \checkmark & ~ & ~ & \checkmark & ~ & p & \checkmark & \checkmark & ~ & \checkmark & ~ \\ 
        Büchel \cite{buechel_ontology-based_2017}& \checkmark & ~ & \checkmark & \checkmark & \checkmark & pv & \checkmark & \checkmark & \checkmark & \checkmark & \checkmark \\ 
        Regele \cite{regele_using_2008}  & \checkmark & ~ & ~ & \checkmark & ~ & pva & ~ & \checkmark & ~ & \checkmark & ~ \\ 
        Bogdoll \cite{bogdoll_one_2022} & ~ & ~ & ~ & ~ & \checkmark & pva & \checkmark & ~ & ~ & ~ & \checkmark \\ 
        Huang \cite{huang_ontology-based_2019}& \checkmark & ~ & \checkmark & \checkmark & \checkmark & pv & \checkmark & ~ & \checkmark & \checkmark & ~ \\ 
        Zhao \cite{zhao_ontology-based_2017} & \checkmark & \checkmark & \checkmark & \checkmark & ~ & pv & \checkmark & ~ & ~ & \checkmark & ~ \\ 
        Fang \cite{fang_ontology-based_2019} & \checkmark & ~ & ~ & \checkmark & \checkmark & pv & \checkmark & ~ & ~ & \checkmark & ~ \\ 
        Kohlhaas \cite{kohlhaas_semantic_2014} & \checkmark & ~ & ~ & ~ & \checkmark & pv & ~ & ~ & ~ & ~ & ~ \\
        Herrmann \cite{herrmann_using_2022} & ~ & \checkmark & ~ & ~ & ~ & p & ~ & ~ & \checkmark & ~ & \checkmark \\ 
        Westhofen \cite{westhofen_using_2022} & \checkmark & ~ & \checkmark & \checkmark & \checkmark & pva & \checkmark & \checkmark & \checkmark & \checkmark & \checkmark \\
        \bottomrule
    \end{tabular}
    \label{tab:categories}
\end{table*}
In addition to the level of detail of the ontology, the concrete entities and relationships are also relevant. These ultimately determine whether a description of the environment is sufficient for the desired use case.
In order to make the different ontologies comparable and to be able to evaluate them, we defined twelve categories. Based on these categories, we examined whether the individual ontologies are able to describe these objects and relations (see \Cref{tab:categories}).
We believe that the road is particularly important in scenario-based testing. Since the road geometry and the associated road topology can vary unlimited, this should be defined particularly detailed in the ontology. Furthermore, the traffic participants' states (State of TP), i.e. positions (p), velocities (v) or accelerations (a) are relevant and how they behave in general in the scenario (actions/intentions).
Moreover, we agree that actions or at least the intentions of the traffic participants should be explicitly specified. A general rule does not always lead to the expected behaviour. For example, a right-of-way rule does not necessarily lead to all traffic participants obeying it at all times.
Rather, the actions in combination with the environment (Static Objects, Signs and Road) are relevant to evaluate the scenario, which can also be shown by the majority of ontologies in \Cref{tab:categories}.
Furthermore, surface conditions of the road or weather are rarely included in the ontologies.

\section{Conclusion}
\label{sec:conclusion}
In this comprehensive review, we provide a detailed overview of ontologies for autonomous driving and classify them into scenario-based validation, as well as situation assessment and decision-making. We discuss the proposed ontologies and research approaches in detail, compare them and outline their contributions to research. Furthermore, we categorize the ontologies into a two-dimensional overview formed by a static and dynamic abstraction axis. This categorization supports in choosing the appropriate ontology for one's validation task along the V-model, which fits to the abstraction level of one's simulation environment. Additionally, we compare the entities and relations of the various ontologies across twelve categories and find out that the road topology, as well as the traffic participants' states and actions are the most relevant entities within an ontology for scenario-based validation.

\section{Acknowledgement}
The research leading to these results is funded by the German Federal Ministry for Economic Affairs and Climate Action" within the project “Verifikations- und Validierungsmethoden automatisierter Fahrzeuge im urbanen Umfeld". The authors would like to thank the consortium for the successful cooperation.

\bibliographystyle{IEEEtran}
\bibliography{bib_options,references,references_nina}

\begin{thebibliography}{10}
\providecommand{\url}[1]{#1}
\csname url@samestyle\endcsname
\providecommand{\newblock}{\relax}
\providecommand{\bibinfo}[2]{#2}
\providecommand{\BIBentrySTDinterwordspacing}{\spaceskip=0pt\relax}
\providecommand{\BIBentryALTinterwordstretchfactor}{4}
\providecommand{\BIBentryALTinterwordspacing}{\spaceskip=\fontdimen2\font plus
\BIBentryALTinterwordstretchfactor\fontdimen3\font minus
  \fontdimen4\font\relax}
\providecommand{\BIBforeignlanguage}[2]{{%
\expandafter\ifx\csname l@#1\endcsname\relax
\typeout{** WARNING: IEEEtran.bst: No hyphenation pattern has been}%
\typeout{** loaded for the language `#1'. Using the pattern for}%
\typeout{** the default language instead.}%
\else
\language=\csname l@#1\endcsname
\fi
#2}}
\providecommand{\BIBdecl}{\relax}
\BIBdecl

\bibitem{scholtes_6-layer_2021}
\BIBentryALTinterwordspacing
M.~Scholtes \emph{et~al.}, ``\BIBforeignlanguage{en}{6-{Layer} {Model} for a
  {Structured} {Description} and {Categorization} of {Urban} {Traffic} and
  {Environment}},'' Feb. 2021, arXiv:2012.06319 [cs]. [Online]. Available:
  \url{http://arxiv.org/abs/2012.06319}
\BIBentrySTDinterwordspacing

\bibitem{ulbrich_defining_2015}
\BIBentryALTinterwordspacing
S.~Ulbrich \emph{et~al.}, ``\BIBforeignlanguage{en}{Defining and
  {Substantiating} the {Terms} {Scene}, {Situation}, and {Scenario} for
  {Automated} {Driving}},'' in \emph{\BIBforeignlanguage{en}{2015 {IEEE} 18th
  {International} {Conference} on {Intelligent} {Transportation}
  {Systems}}}.\hskip 1em plus 0.5em minus 0.4em\relax Gran Canaria, Spain:
  IEEE, Sep. 2015, pp. 982--988. [Online]. Available:
  \url{http://ieeexplore.ieee.org/document/7313256/}
\BIBentrySTDinterwordspacing

\bibitem{stellet_testing_2015}
\BIBentryALTinterwordspacing
J.~E. Stellet \emph{et~al.}, ``\BIBforeignlanguage{en}{Testing of {Advanced}
  {Driver} {Assistance} {Towards} {Automated} {Driving}: {A} {Survey} and
  {Taxonomy} on {Existing} {Approaches} and {Open} {Questions}},'' in
  \emph{\BIBforeignlanguage{en}{2015 {IEEE} 18th {International} {Conference}
  on {Intelligent} {Transportation} {Systems}}}.\hskip 1em plus 0.5em minus
  0.4em\relax Gran Canaria, Spain: IEEE, Sep. 2015, pp. 1455--1462. [Online].
  Available: \url{http://ieeexplore.ieee.org/document/7313330/}
\BIBentrySTDinterwordspacing

\bibitem{feng_intelligent_2021}
\BIBentryALTinterwordspacing
S.~Feng \emph{et~al.}, ``\BIBforeignlanguage{en}{Intelligent driving
  intelligence test for autonomous vehicles with naturalistic and adversarial
  environment},'' \emph{\BIBforeignlanguage{en}{Nature Communications}},
  vol.~12, no.~1, p. 748, Feb. 2021. [Online]. Available:
  \url{https://www.nature.com/articles/s41467-021-21007-8}
\BIBentrySTDinterwordspacing

\bibitem{elgharbawy_ontology-based_2019}
\BIBentryALTinterwordspacing
M.~Elgharbawy \emph{et~al.}, ``\BIBforeignlanguage{en}{Ontology-based adaptive
  testing for automated driving functions using data mining techniques},''
  \emph{\BIBforeignlanguage{en}{Transportation Research Part F: Traffic
  Psychology and Behaviour}}, vol.~66, pp. 234--251, Oct. 2019. [Online].
  Available:
  \url{https://linkinghub.elsevier.com/retrieve/pii/S1369847818306351}
\BIBentrySTDinterwordspacing

\bibitem{ding_cmts_2019}
\BIBentryALTinterwordspacing
W.~Ding \emph{et~al.}, ``\BIBforeignlanguage{en}{{CMTS}: {Conditional}
  {Multiple} {Trajectory} {Synthesizer} for {Generating} {Safety}-critical
  {Driving} {Scenarios}},'' Oct. 2019, arXiv:1910.00099 [cs, eess, stat].
  [Online]. Available: \url{http://arxiv.org/abs/1910.00099}
\BIBentrySTDinterwordspacing

\bibitem{langner_estimating_2018}
\BIBentryALTinterwordspacing
J.~Langner \emph{et~al.}, ``\BIBforeignlanguage{en}{Estimating the {Uniqueness}
  of {Test} {Scenarios} derived from {Recorded} {Real}-{World}-{Driving}-{Data}
  using {Autoencoders}},'' in \emph{\BIBforeignlanguage{en}{2018 {IEEE}
  {Intelligent} {Vehicles} {Symposium} ({IV})}}.\hskip 1em plus 0.5em minus
  0.4em\relax Changshu: IEEE, Jun. 2018, pp. 1860--1866. [Online]. Available:
  \url{https://ieeexplore.ieee.org/document/8500464/}
\BIBentrySTDinterwordspacing

\bibitem{riedmaier_survey_2020}
\BIBentryALTinterwordspacing
S.~Riedmaier \emph{et~al.}, ``\BIBforeignlanguage{en}{Survey on
  {Scenario}-{Based} {Safety} {Assessment} of {Automated} {Vehicles}},''
  \emph{\BIBforeignlanguage{en}{IEEE Access}}, vol.~8, pp. 87\,456--87\,477,
  2020. [Online]. Available:
  \url{https://ieeexplore.ieee.org/document/9090897/}
\BIBentrySTDinterwordspacing

\bibitem{wotawa_ontology-based_2020}
\BIBentryALTinterwordspacing
F.~Wotawa \emph{et~al.}, ``\BIBforeignlanguage{en}{Ontology-based {Testing}:
  {An} {Emerging} {Paradigm} for {Modeling} and {Testing} {Systems} and
  {Software}},'' in \emph{\BIBforeignlanguage{en}{2020 {IEEE} {International}
  {Conference} on {Software} {Testing}, {Verification} and {Validation}
  {Workshops} ({ICSTW})}}.\hskip 1em plus 0.5em minus 0.4em\relax Porto,
  Portugal: IEEE, Oct. 2020, pp. 14--17. [Online]. Available:
  \url{https://ieeexplore.ieee.org/document/9155880/}
\BIBentrySTDinterwordspacing

\bibitem{klueck_using_2018}
\BIBentryALTinterwordspacing
F.~Klueck \emph{et~al.}, ``\BIBforeignlanguage{en}{Using {Ontologies} for
  {Test} {Suites} {Generation} for {Automated} and {Autonomous} {Driving}
  {Functions}},'' in \emph{\BIBforeignlanguage{en}{2018 {IEEE} {International}
  {Symposium} on {Software} {Reliability} {Engineering} {Workshops}
  ({ISSREW})}}.\hskip 1em plus 0.5em minus 0.4em\relax Memphis, TN: IEEE, Oct.
  2018, pp. 118--123. [Online]. Available:
  \url{https://ieeexplore.ieee.org/document/8539174/}
\BIBentrySTDinterwordspacing

\bibitem{leitner_enable-s3testing_2019}
\BIBentryALTinterwordspacing
A.~Leitner \emph{et~al.}, ``{ENABLE}-{S3}:{Testing} \& {Validation} of {Highly}
  {Automated} {Systems},'' 2019. [Online]. Available:
  \url{https://drive.google.com/file/d/15c1Oe69dpvW5dma8-uS8hev17x-6V3zU/view}
\BIBentrySTDinterwordspacing

\bibitem{butz_virtual_2020}
\BIBentryALTinterwordspacing
T.~Butz \emph{et~al.}, ``\BIBforeignlanguage{en}{Virtual {Test} of {Automated}
  {Driving} {Functions}},'' \emph{\BIBforeignlanguage{en}{ATZ worldwide}}, vol.
  122, no.~5, pp. 16--21, May 2020. [Online]. Available:
  \url{https://link.springer.com/10.1007/s38311-020-0228-7}
\BIBentrySTDinterwordspacing

\bibitem{bagschik_ontology_2018}
\BIBentryALTinterwordspacing
G.~Bagschik \emph{et~al.}, ``\BIBforeignlanguage{en}{Ontology based {Scene}
  {Creation} for the {Development} of {Automated} {Vehicles}},''
  \emph{\BIBforeignlanguage{en}{arXiv:1704.01006 [cs]}}, Apr. 2018, arXiv:
  1704.01006. [Online]. Available: \url{http://arxiv.org/abs/1704.01006}
\BIBentrySTDinterwordspacing

\bibitem{chen_ontology-based_2018}
\BIBentryALTinterwordspacing
W.~Chen and L.~Kloul, ``\BIBforeignlanguage{en}{An {Ontology}-based {Approach}
  to {Generate} the {Advanced} {Driver} {Assistance} {Use} {Cases} of {Highway}
  {Traffic}:},'' in \emph{\BIBforeignlanguage{en}{Proceedings of the 10th
  {International} {Joint} {Conference} on {Knowledge} {Discovery}, {Knowledge}
  {Engineering} and {Knowledge} {Management}}}.\hskip 1em plus 0.5em minus
  0.4em\relax Seville, Spain: SCITEPRESS - Science and Technology Publications,
  2018, pp. 75--83. [Online]. Available:
  \url{http://www.scitepress.org/DigitalLibrary/Link.aspx?doi=10.5220/0006931700750083}
\BIBentrySTDinterwordspacing

\bibitem{nalic_scenario_2020}
D.~Nalic \emph{et~al.}, ``\BIBforeignlanguage{en}{Scenario {Based} {Testing} of
  {Automated} {Driving} {Systems}: {A} {Literature} {Survey}},'' p.~11, 2020.

\bibitem{schafer_no_2017}
\BIBentryALTinterwordspacing
F.~Schafer \emph{et~al.}, ``\BIBforeignlanguage{en}{No need to learn from each
  other? — {Potentials} of knowledge modeling in autonomous vehicle systems
  engineering towards new methods in multidisciplinary contexts},'' in
  \emph{\BIBforeignlanguage{en}{2017 {International} {Conference} on
  {Engineering}, {Technology} and {Innovation} ({ICE}/{ITMC})}}.\hskip 1em plus
  0.5em minus 0.4em\relax Funchal: IEEE, Jun. 2017, pp. 462--468. [Online].
  Available: \url{http://ieeexplore.ieee.org/document/8279921/}
\BIBentrySTDinterwordspacing

\bibitem{brunner_ontologies_2017}
\BIBentryALTinterwordspacing
S.~Brunner \emph{et~al.}, ``\BIBforeignlanguage{en}{Ontologies used in
  robotics: {A} survey with an outlook for automated driving},'' in
  \emph{\BIBforeignlanguage{en}{2017 {IEEE} {International} {Conference} on
  {Vehicular} {Electronics} and {Safety} ({ICVES})}}.\hskip 1em plus 0.5em
  minus 0.4em\relax Vienna, Austria: IEEE, Jun. 2017, pp. 81--84. [Online].
  Available: \url{http://ieeexplore.ieee.org/document/7991905/}
\BIBentrySTDinterwordspacing

\bibitem{gasevic_dragan_model_2009}
\BIBentryALTinterwordspacing
{Gasevic, Dragan} \emph{et~al.}, \emph{\BIBforeignlanguage{en}{Model {Driven}
  {Engineering} and {Ontology} {Development}}}.\hskip 1em plus 0.5em minus
  0.4em\relax Berlin, Heidelberg: Springer Berlin Heidelberg, 2009. [Online].
  Available: \url{http://link.springer.com/10.1007/978-3-642-00282-3}
\BIBentrySTDinterwordspacing

\bibitem{guarino_nicola_handbook_2009}
\BIBentryALTinterwordspacing
{Guarino, Nicola}, \emph{\BIBforeignlanguage{en}{Handbook on {Ontologies}}},
  S.~Staab and R.~Studer, Eds.\hskip 1em plus 0.5em minus 0.4em\relax Berlin,
  Heidelberg: Springer Berlin Heidelberg, 2009. [Online]. Available:
  \url{http://link.springer.com/10.1007/978-3-540-92673-3}
\BIBentrySTDinterwordspacing

\bibitem{studer_knowledge_1998}
\BIBentryALTinterwordspacing
R.~Studer \emph{et~al.}, ``\BIBforeignlanguage{en}{Knowledge engineering:
  {Principles} and methods},'' \emph{\BIBforeignlanguage{en}{Data \& Knowledge
  Engineering}}, vol.~25, no. 1-2, pp. 161--197, Mar. 1998. [Online].
  Available:
  \url{https://linkinghub.elsevier.com/retrieve/pii/S0169023X97000566}
\BIBentrySTDinterwordspacing

\bibitem{Genesereth01}
M.~R. Genesereth and N.~J. Nilsson, \emph{{Logical Foundations of Artificial
  Intelligence}}.\hskip 1em plus 0.5em minus 0.4em\relax Los Altos, CA: Morgan
  Kaufmann, 1987.

\bibitem{berners-lee_tim_semantic_2001}
\BIBentryALTinterwordspacing
{Berners-Lee, Tim} \emph{et~al.}, ``The {Semantic} {Web},'' \emph{Scientific
  American, a division of Nature America, Inc.}, no. Scientific American, pp.
  34--43, 2001. [Online]. Available: \url{http://www.jstor.org/stable/26059207}
\BIBentrySTDinterwordspacing

\bibitem{jakus_concepts_2013}
\BIBentryALTinterwordspacing
G.~Jakus \emph{et~al.}, \emph{\BIBforeignlanguage{en}{Concepts, {Ontologies},
  and {Knowledge} {Representation}}}, ser. {SpringerBriefs} in {Computer}
  {Science}.\hskip 1em plus 0.5em minus 0.4em\relax New York, NY: Springer New
  York, 2013. [Online]. Available:
  \url{https://link.springer.com/10.1007/978-1-4614-7822-5}
\BIBentrySTDinterwordspacing

\bibitem{lv_approach_2011}
\BIBentryALTinterwordspacing
Y.~Lv, ``\BIBforeignlanguage{en}{An approach to ontologies integration},'' in
  \emph{\BIBforeignlanguage{en}{2011 {Eighth} {International} {Conference} on
  {Fuzzy} {Systems} and {Knowledge} {Discovery} ({FSKD})}}.\hskip 1em plus
  0.5em minus 0.4em\relax Shanghai: IEEE, Jul. 2011, pp. 1262--1266. [Online].
  Available: \url{http://ieeexplore.ieee.org/document/6019706/}
\BIBentrySTDinterwordspacing

\bibitem{fensel_web_2008}
\BIBentryALTinterwordspacing
``\BIBforeignlanguage{en}{From {Web} to {Semantic} {Web}},'' in
  \emph{\BIBforeignlanguage{en}{Implementing {Semantic} {Web} {Services}}},
  D.~Fensel \emph{et~al.}, Eds.\hskip 1em plus 0.5em minus 0.4em\relax Berlin,
  Heidelberg: Springer Berlin Heidelberg, 2008, pp. 3--25. [Online]. Available:
  \url{http://link.springer.com/10.1007/978-3-540-77020-6_1}
\BIBentrySTDinterwordspacing

\bibitem{menzel_functional_2019}
\BIBentryALTinterwordspacing
T.~Menzel \emph{et~al.}, ``\BIBforeignlanguage{en}{From {Functional} to
  {Logical} {Scenarios}: {Detailing} a {Keyword}-{Based} {Scenario}
  {Description} for {Execution} in a {Simulation} {Environment}},''
  \emph{\BIBforeignlanguage{en}{arXiv:1905.03989 [cs]}}, May 2019, arXiv:
  1905.03989. [Online]. Available: \url{http://arxiv.org/abs/1905.03989}
\BIBentrySTDinterwordspacing

\bibitem{kloul_advanced_2020}
\BIBentryALTinterwordspacing
L.~Kloul and W.~Chen, ``\BIBforeignlanguage{en}{An {Advanced} {Driver}
  {Assistance} {Test} {Cases} {Generation} {Methodology} {Based} on {Highway}
  {Traffic} {Situation} {Description} {Ontologies}},'' in
  \emph{\BIBforeignlanguage{en}{Knowledge {Discovery}, {Knowledge}
  {Engineering} and {Knowledge} {Management}}}, A.~Fred \emph{et~al.},
  Eds.\hskip 1em plus 0.5em minus 0.4em\relax Cham: Springer International
  Publishing, 2020, vol. 1222, pp. 93--113, series Title: Communications in
  Computer and Information Science. [Online]. Available:
  \url{http://link.springer.com/10.1007/978-3-030-49559-6_5}
\BIBentrySTDinterwordspacing

\bibitem{medrano-berumen_abstract_2019}
\BIBentryALTinterwordspacing
C.~Medrano-Berumen and M.~I. Akbas, ``\BIBforeignlanguage{en}{Abstract
  {Simulation} {Scenario} {Generation} for {Autonomous} {Vehicle}
  {Verification}},'' in \emph{\BIBforeignlanguage{en}{2019
  {SoutheastCon}}}.\hskip 1em plus 0.5em minus 0.4em\relax Huntsville, AL, USA:
  IEEE, Apr. 2019, pp. 1--6. [Online]. Available:
  \url{https://ieeexplore.ieee.org/document/9020575/}
\BIBentrySTDinterwordspacing

\bibitem{feilhauer_absicherung_2018}
\BIBentryALTinterwordspacing
M.~C. Feilhauer, ``\BIBforeignlanguage{de}{Absicherung von
  {Fahrerassistenzsystemen}},'' Ph.D. dissertation, Universität Stuttgart,
  Stuttgart, 2018. [Online]. Available:
  \url{https://elib.uni-stuttgart.de/handle/11682/10283}
\BIBentrySTDinterwordspacing

\bibitem{schuldt_effiziente_2013}
\BIBentryALTinterwordspacing
F.~Schuldt, ``\BIBforeignlanguage{de}{Effiziente systematische
  {Testgenerierung} für {Fahrerassistenzsysteme} in virtuellen
  {Umgebungen}},'' 2013. [Online]. Available:
  \url{https://publikationsserver.tu-braunschweig.de/servlets/MCRFileNodeServlet/dbbs_derivate_00031187/AAET_Schuldt_Saust_Lichte_Maurer_Scholz.pdf}
\BIBentrySTDinterwordspacing

\bibitem{herrmann_using_2022}
\BIBentryALTinterwordspacing
M.~Herrmann \emph{et~al.}, ``\BIBforeignlanguage{en}{Using ontologies for
  dataset engineering in automotive {AI} applications},'' in
  \emph{\BIBforeignlanguage{en}{2022 {Design}, {Automation} \& {Test} in
  {Europe} {Conference} \& {Exhibition} ({DATE})}}.\hskip 1em plus 0.5em minus
  0.4em\relax Antwerp, Belgium: IEEE, Mar. 2022, pp. 526--531. [Online].
  Available: \url{https://ieeexplore.ieee.org/document/9774675/}
\BIBentrySTDinterwordspacing

\bibitem{bogdoll_one_2022}
\BIBentryALTinterwordspacing
D.~Bogdoll \emph{et~al.}, ``\BIBforeignlanguage{en}{One {Ontology} to {Rule}
  {Them} {All}: {Corner} {Case} {Scenarios} for {Autonomous} {Driving}},'' Oct.
  2022, arXiv:2209.00342 [cs]. [Online]. Available:
  \url{http://arxiv.org/abs/2209.00342}
\BIBentrySTDinterwordspacing

\bibitem{westhofen_using_2022}
\BIBentryALTinterwordspacing
L.~Westhofen \emph{et~al.}, ``\BIBforeignlanguage{en}{Using {Ontologies} for
  the {Formalization} and {Recognition} of {Criticality} for {Automated}
  {Driving}},'' \emph{\BIBforeignlanguage{en}{IEEE Open Journal of Intelligent
  Transportation Systems}}, vol.~3, pp. 519--538, 2022. [Online]. Available:
  \url{https://ieeexplore.ieee.org/document/9810486/}
\BIBentrySTDinterwordspacing

\bibitem{ulbrich_graph-based_2014}
\BIBentryALTinterwordspacing
S.~Ulbrich \emph{et~al.}, ``\BIBforeignlanguage{en}{Graph-based context
  representation, environment modeling and information aggregation for
  automated driving},'' in \emph{\BIBforeignlanguage{en}{2014 {IEEE}
  {Intelligent} {Vehicles} {Symposium} {Proceedings}}}.\hskip 1em plus 0.5em
  minus 0.4em\relax MI, USA: IEEE, Jun. 2014, pp. 541--547. [Online].
  Available: \url{http://ieeexplore.ieee.org/document/6856556/}
\BIBentrySTDinterwordspacing

\bibitem{hulsen_traffic_2011}
\BIBentryALTinterwordspacing
M.~Hülsen \emph{et~al.}, ``\BIBforeignlanguage{en}{Traffic intersection
  situation description ontology for advanced driver assistance},'' in
  \emph{\BIBforeignlanguage{en}{2011 {IEEE} {Intelligent} {Vehicles}
  {Symposium} ({IV})}}.\hskip 1em plus 0.5em minus 0.4em\relax Baden-Baden,
  Germany: IEEE, Jun. 2011, pp. 993--999. [Online]. Available:
  \url{http://ieeexplore.ieee.org/document/5940415/}
\BIBentrySTDinterwordspacing

\bibitem{buechel_ontology-based_2017}
\BIBentryALTinterwordspacing
M.~Buechel \emph{et~al.}, ``\BIBforeignlanguage{en}{Ontology-based traffic
  scene modeling, traffic regulations dependent situational awareness and
  decision-making for automated vehicles},'' in
  \emph{\BIBforeignlanguage{en}{2017 {IEEE} {Intelligent} {Vehicles}
  {Symposium} ({IV})}}.\hskip 1em plus 0.5em minus 0.4em\relax Los Angeles, CA,
  USA: IEEE, Jun. 2017, pp. 1471--1476. [Online]. Available:
  \url{http://ieeexplore.ieee.org/document/7995917/}
\BIBentrySTDinterwordspacing

\bibitem{regele_using_2008}
\BIBentryALTinterwordspacing
R.~Regele, ``\BIBforeignlanguage{en}{Using {Ontology}-{Based} {Traffic}
  {Models} for {More} {Efficient} {Decision} {Making} of {Autonomous}
  {Vehicles}},'' in \emph{\BIBforeignlanguage{en}{Fourth {International}
  {Conference} on {Autonomic} and {Autonomous} {Systems} ({ICAS}'08)}}.\hskip
  1em plus 0.5em minus 0.4em\relax Gosier, Guadeloupe: IEEE, Mar. 2008, pp.
  94--99. [Online]. Available:
  \url{http://ieeexplore.ieee.org/document/4488328/}
\BIBentrySTDinterwordspacing

\bibitem{fang_ontology-based_2019}
\BIBentryALTinterwordspacing
F.~Fang \emph{et~al.}, ``\BIBforeignlanguage{en}{Ontology-based {Reasoning}
  {Approach} for {Long}-term {Behavior} {Prediction} of {Road} {Users}},'' in
  \emph{\BIBforeignlanguage{en}{2019 {IEEE} {Intelligent} {Transportation}
  {Systems} {Conference} ({ITSC})}}.\hskip 1em plus 0.5em minus 0.4em\relax
  Auckland, New Zealand: IEEE, Oct. 2019, pp. 2068--2073. [Online]. Available:
  \url{https://ieeexplore.ieee.org/document/8917526/}
\BIBentrySTDinterwordspacing

\bibitem{huang_ontology-based_2019}
\BIBentryALTinterwordspacing
L.~Huang \emph{et~al.}, ``\BIBforeignlanguage{en}{Ontology-{Based} {Driving}
  {Scene} {Modeling}, {Situation} {Assessment} and {Decision} {Making} for
  {Autonomous} {Vehicles}},'' in \emph{\BIBforeignlanguage{en}{2019 4th
  {Asia}-{Pacific} {Conference} on {Intelligent} {Robot} {Systems}
  ({ACIRS})}}.\hskip 1em plus 0.5em minus 0.4em\relax Nagoya, Japan: IEEE, Jul.
  2019, pp. 57--62. [Online]. Available:
  \url{https://ieeexplore.ieee.org/document/8935984/}
\BIBentrySTDinterwordspacing

\bibitem{zhao_ontology-based_2017}
\BIBentryALTinterwordspacing
L.~Zhao \emph{et~al.}, ``\BIBforeignlanguage{en}{Ontology-{Based} {Driving}
  {Decision} {Making}: {A} {Feasibility} {Study} at {Uncontrolled}
  {Intersections}},'' \emph{\BIBforeignlanguage{en}{IEICE Transactions on
  Information and Systems}}, vol. E100.D, no.~7, pp. 1425--1439, 2017.
  [Online]. Available:
  \url{https://www.jstage.jst.go.jp/article/transinf/E100.D/7/E100.D_2016EDP7337/_article}
\BIBentrySTDinterwordspacing

\bibitem{kohlhaas_semantic_2014}
\BIBentryALTinterwordspacing
R.~Kohlhaas \emph{et~al.}, ``\BIBforeignlanguage{en}{Semantic state space for
  high-level maneuver planning in structured traffic scenes},'' in
  \emph{\BIBforeignlanguage{en}{17th {International} {IEEE} {Conference} on
  {Intelligent} {Transportation} {Systems} ({ITSC})}}.\hskip 1em plus 0.5em
  minus 0.4em\relax Qingdao, China: IEEE, Oct. 2014, pp. 1060--1065. [Online].
  Available: \url{http://ieeexplore.ieee.org/document/6957828/}
\BIBentrySTDinterwordspacing

\bibitem{deng_skeleton_2018}
Q.~Deng \emph{et~al.}, ``Skeleton model based behavior recognition for
  pedestrians and cyclists from vehicle sce ne camera,'' in \emph{2018 {IEEE}
  {Intelligent} {Vehicles} {Symposium} ({IV})}, Jun. 2018, pp. 1293--1298,
  iSSN: 1931-0587.

\bibitem{asam_asam_nodate}
\BIBentryALTinterwordspacing
ASAM., ``{ASAM} {OpenDRIVE}®.'' [Online]. Available:
  \url{https://www.asam.net/standards/detail/opendrive/}
\BIBentrySTDinterwordspacing

\bibitem{magosi_survey_2022}
\BIBentryALTinterwordspacing
Z.~F. Magosi \emph{et~al.}, ``\BIBforeignlanguage{en}{A {Survey} on {Modelling}
  of {Automotive} {Radar} {Sensors} for {Virtual} {Test} and {Validation} of
  {Automated} {Driving}},'' \emph{\BIBforeignlanguage{en}{Sensors}}, vol.~22,
  no.~15, p. 5693, Jul. 2022. [Online]. Available:
  \url{https://www.mdpi.com/1424-8220/22/15/5693}
\BIBentrySTDinterwordspacing

\bibitem{schuldt_beitrag_2017}
F.~Schuldt, ``Ein {Beitrag} für den methodischen {Test} von automatisierten
  {Fahrfunktionen} mit {Hilfe} von virtuellen {Umgebungen},'' 2017.

\end{thebibliography}
\end{document}